%% file: main.tex
\documentclass[conference]{IEEEtran}  
\IEEEoverridecommandlockouts
\usepackage{cite}
\usepackage{amsmath,amsfonts,amsthm,bm} % Math packages
\usepackage{caption}
\usepackage{subcaption} %To insert subfigures 
\usepackage{verbatim}  %To comment out large sections of text with \begin{comment} and \end{comment}
\usepackage{colortbl} %For using colors in header of tables
\usepackage{algorithm}
\usepackage{algpseudocode}
\usepackage{graphicx}
\usepackage{textcomp}
\usepackage[table]{xcolor}

\usepackage{array,multirow} %For changing borders in tables

\usepackage{tabularx,booktabs}
\newcolumntype{C}{>{\centering\arraybackslash}X} % centered version of "X" type
\definecolor{maroon}{cmyk}{0,0.87,0.68,0.32} %For headers in table instead of gray (it has different intensities, for example: !10 has less intensity)
\usepackage{flushend} %Align last page (same length both columns)
\usepackage{url} 
\usepackage{authblk} %To add authors from different departments and universities

\usepackage{lipsum}  %Generate dummy text
\usepackage{amssymb} % To use symbol greater or equal approximately: \gtrapprox

\usepackage{mathtools}

\DeclarePairedDelimiter\floor{\lfloor}{\rfloor}

\newcommand{\Psib}{\mathbf{\Psi}}
\newcommand{\xb}{\mathbf{x}}
\newcommand{\yb}{\mathbf{y}}
\newcommand{\zb}{\mathbf{z}}
\newcommand{\fb}{\mathbf{f}}

\newcommand{\wb}{\mathbf{w}}

\newcommand{\Phib}{\mathbf{\Phi}}
\newcommand{\Ab}{\mathbf{A}}

\newcommand{\M}{M}
\newcommand{\N}{N}

\def\BibTeX{{\rm B\kern-.05em{\sc i\kern-.025em b}\kern-.08em
    T\kern-.1667em\lower.7ex\hbox{E}\kern-.125emX}}
\begin{document}

% \title{Sparse recovery of compressive sensed data in hyperspectral imaging}
\title{A Comparative Study of Compressive Sensing Algorithms for Hyperspectral Imaging Reconstruction}

% \author{
% \IEEEauthorblockN{Jon Alvarez Justo, Daniela Lupu, Milica Orlandić, Ion Necoară, Tor Arne Johansen}
% \IEEEauthorblockA{\textit{ \textcolor{blue}{Add departments: Automatic Control and Systems Engineering}} \\
% \textcolor{blue}{\textit{Add universities (NTNU, Politehnica University of Bucharest)}}\\
% \textcolor{blue}{Add locations (Trondheim, Norway - Bucharest, Romania)} \\
% \textcolor{blue}{Add emails}}
% }
   
\author[1]{Jon Alvarez Justo}%\thanks{jon.a.justo@ntnu.no}}
\author[2]{Daniela Lupu}%\thanks{daniela.lupu@upb.ro}}
\author[1]{Milica Orlandi{\'c}}%\thanks{milica.orlandic@ntnu.no}}
\author[2]{Ion Necoara}%\thanks{ion.necoara@upb.ro}}
\author[3]{Tor Arne Johansen}%\thanks{tor.arne.johansen@ntnu.no}}

\affil[1]{Department of Electronic Systems, Norwegian University of Science and Technology }
\affil[2]{Department of Automatic Control and Systems Engineering,  University Politehnica Bucharest }
\affil[3]{Department of Engineering Cybernetics, Norwegian University of Science and Technology }

\renewcommand\Authands{, and }

\maketitle

\input{0-Abstract_index_terms}

\input{1-Introduction}

\input{2-Background}

\input{2.1-SRA_intro}

\input{2.2-FISTA_algorithm}

\input{2.3-ADMM_algorithm}

% \input{2.4-CD_algorithm}

\input{2.5-GOMP}

\input{2.6-BIHT}

\input{2.7-CoSaMP}

\input{2.8-Metrics}
\input{4.1-Results_Acc}

\input{4.2-Results_Perf}
\input{5-Conclusions}

\input{6-Acknowledgement}

%\clearpage
\bibliographystyle{unsrt} %Sort bibliography according to order in citations
% % \bibliographystyle{plain}
\bibliography{Bibliography}

\end{document}

%% file: 0-Abstract_index_terms.tex
\begin{abstract}

Hyperspectral Imaging comprises excessive data consequently leading to significant challenges for data processing, storage and transmission. Compressive Sensing has been used in the field of Hyperspectral Imaging as a technique to compress the large amount of data. This work addresses the recovery of hyperspectral images $2.5\times$ compressed. A comparative study in terms of the accuracy and the performance of the convex FISTA/ADMM in addition to the greedy gOMP/BIHT/CoSaMP recovery algorithms is presented. The results indicate that the algorithms recover successfully the compressed data, yet the gOMP algorithm achieves superior accuracy and faster recovery in comparison to the other algorithms at the expense of high dependence on unknown sparsity level of the data to recover.

% \textcolor{magenta}{Structure abstract and conclusion:Abstract (1 paragraph): go through the paper and comment the motivation of the problem to solve and why it is solved that way, methodology proposed, brief mention to the most important conclusion or result, remaining conclusion analysis is for the conclusion section itself.
% Conclusion (1 or 2 paragraph): short summary of what it was done, then summarise main key findings from the results more extensively than in abstract but it's concisely, say implication of this in the field and set further lines of work for it}

\end{abstract}


%% file: 1-Introduction.tex
% STATUS
%   -Finish greedy algorithms (this will be my next step now that there was progress in the results section)
%   
%   -Check pseudo FISTA/ADMM
%   -BIHT 
%   -Add references

\section{Introduction}

\footnotetext{IEEE-copyrighted material - © 2022 IEEE.  Personal use of this material is permitted.  Permission from IEEE must be obtained for all other uses, in any current or future media, including reprinting/republishing this material for advertising or promotional purposes, creating new collective works, for resale or redistribution to servers or lists, or reuse of any copyrighted component of this work in other works.}

Hyperspectral imaging (HSI) collects and processes light from a large number of bands in the electromagnetic spectrum. The resulting images are stacked in data cubes with spatial dimensions $X$ and $Y$, where each pixel has $N$ spectral bands. Platforms with HSI equipment usually have several constraints such as limited storage, windowed transmission times and limited bandwidth in communication data links, which pose a significant challenge for HSI processing due to the excessive amount of data. Consequently, the state of the art~\cite{orlandic2019parallel}~\cite{dua2020comprehensive} proposes a variety of compression techniques to reduce the size of the HSI data, such as CCSDS-123 compression algorithm used in space-related applications and \textit{Compressive Sensing} (CS)~\cite{stankovic2015multimedia}. In CS, the compression is performed by lowering the amount of data either by utilizing a dedicated compressive HSI sensor such as in the  Miniature Ultra-Spectral Imaging (MUSI) system~\cite{august2016miniature} or by performing the subsampling of the data acquired by a regular non-compressive HSI sensor~\cite{august2013compressive}. The subsequent data reconstruction is accomplished by the lossy recovery algorithms classified in \textit{convex} and \textit{greedy}. Examples of convex algorithms are the \textit{Fast Iterative Shrinkage/Thresholding Algorithm} (FISTA)~\cite{beck2009fast}, the \textit{Alternating Direction Method of Multipliers} (ADMM)~\cite{boyd2011distributed}, the \textit{Gradient Descent} (GD)~\cite{bib:original_paper_gradient_descend}, and the \textit{Basis Pursuit} (BP)~\cite{bib:original_paper_BP}. A common pursuit greedy method is the \textit{Orthogonal Matching Pursuit} (OMP)~\cite{bib:original_paper_OMP} which constitutes the basis of more advanced and improved pursuit techniques such as the \textit{Generalized Orthogonal Matching Pursuit} (gOMP)~\cite{bib:original_paper_GOMP} and the \textit{Compressive Sampling Matching Pursuit} (CoSaMP)~\cite{bib:original_paper_CoSaMP}. Examples of thresholding greedy methods are the \textit{Iterative Hard Thresholding} (IHT)~\cite{bib:original_paper_IHT} and its variant called the \textit{Backtracking Iterative Hard Thresholding} (BIHT)~\cite{bib:original_paper_BIHT}. This work performs a comparison of some of these reconstruction algorithms in HSI data in terms of the recovery accuracy as well as the performance analysed through the algorithm convergence, the recovery time and the time scalability. According to the authors knowledge, this paper is the first work comparing specifically the FISTA/ADMM/gOMP/BIHT/CoSaMP algorithms in HSI data, where the BIHT algorithm has not been used in the HSI field so far. 

The remainder of the paper is organized as follows. Section~\ref{Section: background} introduces CS in the context of HSI and describes the recovery algorithms. Section~\ref{Section: results} presents the reconstruction accuracy and the performance results. Finally, Section~\ref{Section: conclusions} concludes the work.

%% file: 2-Background.tex
\section{Background}

\label{Section: background}

Pixel-wise HSI data processing in pushbroom imaging, where a frame of pixels is scanned in a time instance, does not require the acquisition of the complete cube. A spatial pixel $\mathbf{f} \in \mathbb{R}^{N\times1}$ with $N$ spectral samples is represented with the transform equation $ \fb=\Psib\xb$ where $\mathbf{\Psi} \in \mathbb{C}^{N\times N}$ is a \textit{Discrete Fourier Transform} (DFT) matrix and hence $\fb$ and $\mathbf{x} \in \mathbb{C}^{N\times1}$ are the pixel elements expressed respectively in the DFT and the IDFT domains. The pixel $\fb$ is sub-sampled by computing its projections over the randomized measurement matrix $\mathbf{\Phi} \in \mathbb{R}^{M\times N}$ resulting in the measurement vector $\mathbf{y} \in \mathbb{R}^{M\times1}$ with $\M$ ($M$ $<$ $N$) random spectral samples from $\mathbf{f}$: 
\begin{equation}
    \yb=\Phib\fb=  \Phib \Psib \xb = \Ab \xb,
    \label{eq:main_CS_eq}
\end{equation}
\noindent where $\mathbf{A} \in \mathbb{C}^{M\times N}$ is referred as the \textit{dictionary}. Since $\yb = \Ab \xb$ is an under-determined linear system of $\N$ unknowns and $\M$ equations with infinitely many solutions, a sparsity regularization condition is imposed over $\xb$ to enforce  a unique solution to the system. The sparsity of the vector $\xb$ is quantified using the sparsity level $\kappa\ll N$, which gives the number of non-zero samples in $\xb$, i.e., $ \kappa\;=\;{\parallel \mathbf{x} \parallel}_0$. HSI data are not strictly sparse but \textit{compressible}~\cite{draganic2017some} and hence this work employs a pre-processing stage to ensure that the data are $\kappa$-sparse. Namely, a \textit{sparsification} stage is performed in the IDFT domain and hence this domain transformation leads to $\mathbf{f} \in \mathbb{C}^{N\times1}$ and $\mathbf{y} \in \mathbb{C}^{M\times1}$. The samples in $\mathbf{x}$ above a threshold are maintained whereas remaining are rounded to zero when the condition $\mid$$\mathbf{x}^i$$\mid$$\;-\;\mu_x<\;T\cdot\sigma_{x}$ for $i=1,...,N$ is satisfied, where $\mu_x$ and $\sigma_{x}$ $\in \mathbb{R}^{1\times1}$ are respectively the average and the standard deviation of the transform vector $\mid$$\mathbf{x}$$\mid$, and the sparsification factor $T$ adjusts experimentally the sparsity level $\kappa$ of $\mathbf{x}$. For higher $T$ values, fewer non-zero samples are maintained above the threshold and thus a sparser pixel $\xb$ is obtained.

% when condition $\xb\;<\;T\cdot\sigma_{x}$ is satisfied, where $\sigma_{x}$ is the standard deviation of demeaned vector $\mathbf{x}$, and sparsification factor $T$ adjusts experimentally sparsity level $\kappa$ of vector $\mathbf{x}$. For lower $T$ value, fewer non-zero coefficients are maintained above threshold and thus sparser pixel $\xb$ is obtained.

 %$\xb\;-\;\mu_{x}\;<\;T\cdot\sigma_{x}$ 

%% file: 2.1-SRA_intro.tex
\subsection{Sparse Recovery Algorithms}
Optimization algorithms approximate the unique solution to $\mathbf{x}$ in Eq.~(\ref{eq:main_CS_eq}) using not only the measurements in $\mathbf{y}$ and the dictionary $\mathbf{A}$, but also some parameters to adjust the sparsity of $\mathbf{x}$. These algorithms achieve higher accuracy by reducing iteratively the residual $\mathbf{r}$ from the vector $\mathbf{y}$ to a zero vector $\mathbf{0}$. Approximate optimality  is reached when the residual difference $\Delta$ between two consecutive  iterations is below some error tolerance $\epsilon$, or if not achieved, the algorithms stop when  the maximum convergence time  $t_{conv.}$ is reached.

%% file: 2.2-FISTA_algorithm.tex
\subsubsection{Convex Algorithms}

The optimization problem known as \textit{Least Absolute Shrinkage and Selection Operator} - Lasso is given next \cite{tibshirani1996regression}:
\begin{equation} \label{eq:optPb}
    \min_{\xb} H(\xb):=\underbrace{\frac{1}{2}\| \Ab \xb - \yb \|^2}_{:=h(\xb)} + \underbrace{\lambda \| \xb\|_1}_{:=g(\xb)},
\end{equation}
\noindent where the second term of the objective function comprises the regularization parameter $\lambda$ $\in \mathbb{R}^{1\times1}$ which controls the sparsity level of $\xb$ in the $l_1$-norm and needs to be set a priori, and $h(\xb)$ has \textit{Lipschitz gradient} with constant $L = \lambda_{\text{max}}(\Ab^* \Ab)$, i.e., the $L$ is the maximum eigenvalue of the matrix $\Ab^* \Ab$ where $*$ denotes the conjugate transpose. At the same time, $H(\xb)$ is a convex function but non-smooth due to the $l_1$ regularization term. This problem can be solved for example by, a gradient-based method called FISTA presented in Algorithm~\ref{alg:fista} or ADMM shown in Algorithm~\ref{alg:admm}. 

\paragraph{FISTA Algorithm} The updates consist of  a gradient step of $h$ evaluated at the current point (Step 4), followed by a soft-threshold step~\cite{beck2009fast}. %It has  \textcolor{olive}{convergence rate of order $O(1/i^2)$~\cite{beck2009fast}}. 

\begin{algorithm}
	\caption{FISTA}
	\begin{algorithmic}
		\State \textbf{Input:} $\mathbf{y}$, $\mathbf{A}$, $\kappa$, starting point $\xb^0\in \mathbb{R}^{N\times 1}$, Lipschitz constant $L$, $\lambda$, $\epsilon$
		\State \textbf{Initialization:} $t^0=1$, $\zb^0 = \xb^0$,$\mathbf{r}^0 = \yb$, $\Delta=1$ 
		\While {$\Delta \geq \epsilon$}
			\State 1. $\text{aux} \gets \zb^{i-1} - \frac{1}{L} \Ab^*( \Ab \zb^{i-1} - \yb )$
			\State 2. $ \xb^{i} \gets \text{sign}(\text{aux}) \odot \text{max}(|\text{aux}| - \frac{\lambda}{L}, 0)$
			\State 3. $ t^{i} \gets \frac{1+\sqrt{1+4(t^{i-1})^2}}{2}$
		    \State 4. $ \zb^{i} \gets \xb^i + \left( \frac{t^{i-1} - 1}{t^{i}} \right) (\xb^{i} - \xb^{i-1}) $
			\State 5. $ \mathbf{r}^{i} = \yb - \Ab\xb^{i},\,\, \Delta \gets \| \mathbf{r}^{i} - \mathbf{r}^{i-1}\|_2$\\
			\hspace{0.50cm}where $\odot$ is the Hadamard product
			\EndWhile
	\end{algorithmic}
	\label{alg:fista}
\end{algorithm}

%%ORIGINAL PSEUDOCODE FROM DANIELA BELOW

% \begin{algorithm}
% 	\caption{FISTA}
% 	\begin{algorithmic}
% 		\State \textbf{Input:} Starting point $\xb^0\in \mathbb{R}^N $, $L$ Lipschitz constant of $h(\xb)$, $\lambda$, tolerance $\epsilon$
% 		\State \textbf{Initialization:} $\Delta$ =1, $i=1$, $t^0=1$, $\zb^0 = \xb^0$,$\mathbf{r}^0 = \yb$
% 		\While {$\Delta \geq \epsilon$}
% 			\State 1. $\text{aux} \gets \zb^{i-1} - \frac{1}{L} \Ab^T( \Ab \zb^{i-1} - \yb )$
% 			\State 2. $ \xb^{i} \gets \text{sign}(\text{aux}) * \text{max}(|\text{aux}| - \frac{\lambda}{L}, 0)$
% 			\State 3. $ t^{i} \gets \frac{1+\sqrt{1+4(t^{i-1})^2}}{2}$
% 		    \State 4. $ \zb^{i} \gets \xb^i + \left( \frac{t^{i-1} - 1}{t^{i}} \right) (\xb^{i} - \xb^{i-1}) $
% 			\State    $\quad \mathbf{r}^{i} = \yb - \Ab\xb^{i},\,\, \Delta \gets \| \mathbf{r}^{i} - \mathbf{r}^{i-1}\|_2, \, i \gets i+1,$\\
% 			\hspace{0.50cm}where $*$ is the Hadamard product
% 			\EndWhile
% 	\end{algorithmic}
% 	\label{alg:fista}
% \end{algorithm}

%% file: 2.3-ADMM_algorithm.tex
\paragraph{ADMM Algorithm} The Lasso problem reformulated as a constrained problem is solved by introducing a new variable $\zb$ in the function $g$ and imposing the restriction that $\xb - \zb = \mathbf{0}$. The ADMM approach breaks down the problem into two smaller subproblems that are  easier to handle by combining two strategies, namely, the \textit{dual decomposition} and the \textit{augmented Lagrangian methods} for constrained optimization. Thus, the augmented Lagrangian is constructed by moving the constraint in the objective function via a Lagrangian multiplier $\wb$ and a quadratic penalty term for the equality constraints with penalty parameter $\alpha > 0$. This condition ensures that the matrix $\Ab^{*} \Ab+\alpha I$ in Step 1 is invertible. The first two steps of the algorithm entails the minimization of the augmented Lagrangian function with respect to $\xb$ and $\zb$, Step 3 being the dual update~\cite{boyd2011distributed}. %\textcolor{olive}{The convergence rate of ADMM is $O(1/i)$~\cite{boyd2011distributed}}.

\begin{algorithm}
	\caption{ADMM}
	\begin{algorithmic}
		\State \textbf{Input:} $\mathbf{y}$, $\mathbf{A}$, $\kappa$, starting point $\xb^0\in \mathbb{R}^{N\times 1}$, $\lambda$, $\alpha$, $\epsilon$
		\State \textbf{Initialization:} $\zb^0 = \xb^0$, $\wb^0 = \xb^0$,$\mathbf{r}^0 = \yb$, $\Delta=1$ 
		\While{$\Delta \geq \epsilon$}
			\State 1. $\xb^{i} \gets \left(\Ab^{*} \Ab+\alpha I\right)^{-1}\left(\Ab^{*} \yb +\alpha \left(\zb^{i-1}-\wb^{i-1}\right)\right)$
			\State 2. $\zb^{i} \gets \text{sign}(\xb^{i} + \wb^{i-1})\odot\text{max}(|\xb^{i} + \wb^{i-1}|- \lambda \alpha, 0)$
			\State 3. $\wb^{i} \gets \wb^{i-1} + \xb^{i} - \zb^{i}$
			\State 4. $\mathbf{r}^{i} = \yb - \Ab\xb^{i},\,\, \Delta \gets \| \mathbf{r}^{i} - \mathbf{r}^{i-1}\|_2$\\
			\hspace{0.50cm}where $\odot$ is the Hadamard product
			\EndWhile
	\end{algorithmic}
	\label{alg:admm}
\end{algorithm}

%%ORIGINAL PSEUDOCODE FROM DANIELA BELOW

% \begin{algorithm}
% 	\caption{ADMM}
% 	\begin{algorithmic}
% 		\State \textbf{Input:} Starting point $\xb^0\in \mathbb{R}^N $, $\lambda$, $\alpha$, tolerance $\epsilon$.
% 		\State \textbf{Initialization:} $\Delta =1$, $i=1$, $\zb^0 = \xb^0$, $\wb^0 = \xb^0$,$\mathbf{r}^0 = \yb$
% 		\While{$\Delta \geq \epsilon$}
% 			\State 1. $\xb^{i} \gets \left(\Ab^{T} \Ab+\alpha I\right)^{-1}\left(\Ab^{T} \yb +\alpha \left(\zb^{i-1}-\wb^{i-1}\right)\right)$
% 			\State 2. $\zb^{i} \gets \text{sign}(\xb^{i} + \wb^{i-1})*\text{max}(|\xb^{i} + \wb^{i-1}|- \lambda \alpha, 0)$
% 			\State 3. $\wb^{i} \gets \wb^{i-1} + \xb^{i} - \zb^{i}$
% 			\State  $\quad \mathbf{r}^{i} = \yb - \Ab\xb^{i},\,\, \Delta \gets \| \mathbf{r}^{i} - \mathbf{r}^{i-1}\|_2, \, i \gets i+1,$\\
% 			\hspace{0.50cm}where $*$ is the Hadamard product
% 			\EndWhile
% 	\end{algorithmic}
% 	\label{alg:admm}
% \end{algorithm}

%% file: 2.5-GOMP.tex
\subsubsection{Greedy Algorithms} The unique solution to $\mathbf{x}$ is approximated by first calculating the indexes where the significant samples are located in $\mathbf{x}$, and then estimating their sample values.

\paragraph{gOMP Algorithm}
It starts, as shown in Algorithm \ref{alg: GOMP}, computing the indexes of the new significant samples to be calculated in $\mathbf{x}$. The vectors with the indexes are named as \textit{supports}, and the first support vector calculated is $\Theta^i$ which comprises the indexes of the new significant samples in $\mathbf{x}$ to compute, and it is calculated through the $G$ maximum values in magnitude in the vector $\mathbf{p}$ which has the atoms, i.e., the columns in the dictionary $\mathbf{A}$, projected over the residual from the previous iteration. The indexes of significant samples calculated across the iterations are gathered in the support $\mathbf{c}^i$ used for building the sub-dictionary $\mathbf{B}^i$ which contains only the selected atoms in $\mathbf{A}$ which present the maximum projections over the residuals along the iterations, and it allows the approximation of the significant samples in $\mathbf{s}^i$. The samples in $\mathbf{s}^i$ are then assigned to the indexes of $\mathbf{x}$ given in $\mathbf{c}^i$. The greedy algorithms, similar to $\lambda$ in the FISTA/ADMM, demand that the sparsity parameter $\kappa$ of $\mathbf{x}$ needs to be set arbitrarily when the sparsity level is unknown. This parameter is used to ensure the $\kappa$-sparsity of $\mathbf{x}$ by taking only the indexes, given in $\mathbf{q}^i$, of the $\kappa$ maximum samples, where $\mathbf{q}^i$ is used to update the samples in $\mathbf{s}^i$ before building the final $\kappa$-sparse $\mathbf{x}$~\cite{bib:original_paper_GOMP}.

\begin{algorithm}[htbp!]
\caption{gOMP}
	\begin{algorithmic}
	\State \textbf{Input:} $\mathbf{y}$, $\mathbf{A}$, $\kappa$,  $G$, $\epsilon$
	\State \textbf{Output:} $\mathbf{x}$ 
	\State \textbf{Initialization:} ,
	$\mathbf{r}^0=\mathbf{y}$, $\mathbf{c}^{0}=\mathbf{\varnothing}$, $\Delta=1$ 
	
    %  while (norm(r) > tolerance && i < min(K, floor(size(A,1)/S)))
		\While{$\Delta \geq \epsilon$}

		\State 1. $\mathbf{p} \gets \mathbf{A}^{*}\mathbf{r}^{i-1}$, \hspace{0.10cm} $\mathbf{\Theta}^{i} \gets$ argmax$_{G}\left(
                    \mathbf{\mid p\mid}\right)$ 
                    
        \hspace{0.15cm}$\mathbf{c}^{i}\gets \mathbf{\Theta}^{i}
        \cup \mathbf{c}^{i-1}$

	\State 2. $\mathbf{B}^i=\mathbf{A}(\forall,\mathbf{c}^{i})$, 
	$\mathbf{s}^i\gets {\mathbf{B}^i}^{+}\mathbf{y}$

	\State 3. $\mathbf{x} \gets \mathbf{0}$, $\mathbf{x}(\mathbf{c}^{i}) \gets {\mathbf{s}}^{i}$
     $\mathbf{q}^{i}$=argmax$_{\kappa}\left(\mid\mathbf{x}\mid\right)$

     \hspace{0.15cm}$\mathbf{B}^i=\mathbf{A}(\forall,\mathbf{q}^{i})$, $\mathbf{s}^i\gets {\mathbf{B}^i}^{+}\mathbf{y}$
     
	\hspace{0.2cm}$\mathbf{x} \gets \mathbf{0}$, $\mathbf{x}(\mathbf{q}^{i}) \gets \mathbf{s}^{i}$
	
	\State 4. $\mathbf{r}^{i}\gets\mathbf{y}-\mathbf{A}\mathbf{x}$, $\quad \Delta \gets \|\mathbf{r}^{i}-\mathbf{r}^{i-1}\|_2$

	\EndWhile

	%This is used to resize parenthesis:
	%$\left(A+B\right)^{-1}$
    
	\end{algorithmic}
	\label{alg: GOMP}
\end{algorithm}

%% file: 2.6-BIHT.tex
\paragraph{BIHT Algorithm}

Algorithm \ref{alg: BIHT} shows that the main difference of this thresholding algorithm with respect to the other greedy algorithms is that the atom projections are not calculated in Step 1, but the vector $\mathbf{u}$ is used instead as an approximation of vector $\mathbf{x}$ using $\mu$ as constant descent factor~\cite{bib:original_paper_BIHT}.

% Find  i \in S(x) ={ i \in \mathbb{Z}_{+}|  x_i \neq 0 } ? 

\begin{algorithm}[htbp!]
\caption{BIHT}
	\begin{algorithmic}
	\State \textbf{Input:} $\mathbf{y}$, $\mathbf{A}$, $\kappa$, $\mu$, ~$\epsilon$
	\State \textbf{Output:} $\mathbf{x}$

	\State \textbf{Initialization:}  $\mathbf{x}_{a}^0=\mathbf{0}$, $\mathbf{r}^0=\mathbf{y}$, $\Delta=1$ 
	
	\While{$\Delta \geq \epsilon$}

	\State 1. $\mathbf{u}^{i} \gets \mathbf{x}_{a}^{i-1}+\mu \mathbf{A}^{*}\left(\mathbf{y}-\mathbf{A}\mathbf{x}_{a}^{i-1}\right)$, 
	
	\hspace{0.25cm}$\mathbf{\Theta}^{i} \gets$
            argmax$_{\kappa}\left(\mid\mathbf{u}^{i}\mid\right)$, $\mathbf{c}^{i}$=supp$(\mathbf{x}_{a}^{i-1}$), 
            
            \noindent \hspace{0.85cm} $\mathbf{c}^{i}\gets \mathbf{\Theta}^{i} \cup \mathbf{c}^{i}$

    \State 2. $\mathbf{B}^i=\mathbf{A}(\forall, \mathbf{c}^{i})$, $\mathbf{s}^i\gets {\mathbf{B}^i}^{+}\mathbf{y}$

    \State 3. $\mathbf{h}=$argmax$_{\kappa}(\mid\mathbf{s}^{i}\mid)$, $\mathbf{s}^{i}(1, \mathbb{Z} \not \subset \mathbf{h})=\mathbf{0}$, $\mathbf{x}_{\mathbf{a}}^{i}(\mathbf{c}^i)$ $\gets$ $\mathbf{s}^{i}$,
    
    \noindent \hspace{0.8cm} $\mathbf{x}=\mathbf{0}, \mathbf{x}(\mathbf{c}^i)=\mathbf{s}^i$

    \State 4. $\mathbf{r}^{i}\gets\mathbf{y}-\mathbf{A}\mathbf{x}$, $\quad \Delta \gets \|\mathbf{r}^{i}-\mathbf{r}^{i-1}\|_2$

	\EndWhile

	%This is used to resize parenthesis:
	%$\left(A+B\right)^{-1}$
	
	\end{algorithmic}
	\label{alg: BIHT}
\end{algorithm}

%% file: 2.7-CoSaMP.tex
\paragraph{CoSaMP Algorithm} 

Steps 1 and 2 in Algorithm \ref{alg: CoSaMP} are similar to the gOMP algorithm, yet the CoSaMP fixes the number of atoms selected per iteration to $2\kappa$. The CoSaMP addresses step 3 to calculate the $\kappa$-sparse $\mathbf{x}$ by choosing also the $\kappa$ most significant samples as in the gOMP, but already building the final $\mathbf{x}$ using these values instead of performing further computations as it occurs in the gOMP.

\begin{algorithm}[htbp!]
\caption{CoSaMP}
	\begin{algorithmic}
		\State \textbf{Input:} $\mathbf{y}$, $\mathbf{A}$, $\kappa$, $\epsilon$
	\State \textbf{Output:} $\mathbf{x}$
	\State \textbf{Initialization:} $\mathbf{r}^0=\mathbf{y}$, $\mathbf{c}^0=\varnothing$, $\Delta=1$ 
	
	% while (i <= maxiterations) && (norm_error > tolerance);  norm_error=norm(r)/norm(y);
		\While{$\Delta \geq \epsilon$}

    	\State 1.
    	$\mathbf{p}=\mathbf{A}^{*}\mathbf{r}^{i-1}$, $\mathbf{\Theta}^{i} \gets$
            argmax$_{2\kappa}\left(\mid\mathbf{p}\mid\right)$,
                    
       \hspace{0.10cm} $ \mathbf{c}^{i} \gets \mathbf{\Theta}^{i} \cup \mathbf{c}^{i-1} $ 
	
	\State 2. $\mathbf{B}^i=A(\forall,\mathbf{c}^{i})$, $\mathbf{s}^i\gets {\mathbf{B}^i}^{+}\mathbf{y}$
	
	\State 3. $\mathbf{q}^{i} \gets$
            argmax$_{\kappa}\left(\mid\mathbf{s}^i    \mid\right)$, 
                $\mathbf{s}^i\gets \mathbf{s}^i(\mathbf{q}^i)$,
                $\mathbf{c}^i\gets \mathbf{c}^i(\mathbf{q}^i)$

               \hspace{0.2cm}$\mathbf{x}=\mathbf{0}$,
                $\mathbf{x}(\mathbf{c}^i)=\mathbf{s}^i$
              \State 4.  $\mathbf{r}^{i}\gets\mathbf{y}-\mathbf{A}\mathbf{x}$, $\quad \Delta \gets \|\mathbf{r}^{i}-\mathbf{r}^{i-1}\|_2$           
    \EndWhile
    
	\end{algorithmic}
	\label{alg: CoSaMP}
\end{algorithm}

%% file: 4.1-Results_Acc.tex
\section{Results \& Analysis}
\label{Section: results}

\begin{table*}[htbp!]
 \caption{Recovery of Compressed HSI Data Cubes : Accuracy and Performance}
\label{summary_table}
\begin{tabularx}{\textwidth}{@{}l*{5}{C}c@{}}

% \toprule
% Sparsification &  $\mathit{PSNR} \;[\mathrm{dB}]$ &  & &  &   &  &    \\ %\addlinespace

% \midrule
% Salinas  &  $-$ \\ 

% \midrule
% Jasper Ridge &  $-$ \\

% \midrule
% China &  $-$ \\

\toprule
% \textbf{Data Set ($CR=2.5:1$)} & \textbf{Algorithm} & \textbf{Sparsity Parameter}  & $\mathit{PSNR}\;[dB]$  & \textbf{Rec. Time~[s]} & \textbf{Non-Convergence Ratio~[\%]} & \textbf{Convergenvce Iterations}~[K] \\%\addlinespace
% Data Set & Algorithm & Aimed Sparsity   & $\mathit{PSNR}\;[\mathrm{dB}]$  & Recovery Time~[s] & Convergence~[\%] & Iterations.~[K] \\%\addlinespace

Data Set & Algorithm & Aimed Sparsity   & $\mathit{PSNR}\;[\mathrm{dB}]$  &  Iterations~[K]  & Convergence~[\%] & Recovery Time~[s] \\%\addlinespace

\midrule

\multirow{12}{*}{Salinas ($512\times217\times224$)}

 & FISTA  & $\lambda=0.1$  & $40.80$ & $1164700$  &  $84.20$ &  $95645$ \\ 

 &   & $\lambda=100$  & $44.15$ &  $129649$  & $98.87$ & $22423$\\ \cline{2-7}

 & ADMM  &  $\lambda=0.1$ &  $42.96$ & $756202$  & $93.55$ & $82977$ \\ 

 &   & $\lambda=100$  & $44.15$ & $492717$  & $97.19$ & $53360$ \\ \cline{2-7}

 & gOMP  & $\kappa=27$ & $41.85$ & $926$   & $100$  & $654.61$\\

 &   & $\kappa=33$ & $\mathbf{54.67}$ & $943$  & $100$ & $2716$ \\ \cline{2-7}

 & BIHT  & $\kappa=27$ & $42.63$ & $545$  & $99.99$ & $625.19$\\ 

 &   & $\kappa=33$  & $39.52$ & $614$  & $99.93$  & $1751.7$\\  \cline{2-7}

 & CoSaMP  & $\kappa=27$  & $41.15$ & $1351$  & $99.47$ & $2382.7$\\ 

 &   & $\kappa=33$ & $\mathbf{70.98}$ & $2004$  & $100$ & $6189.3$\\ 

\hline
%\addlinespace
\midrule

\multirow{12}{*}{Jasper R. ($100\times100\times198$)}

 & FISTA  & $\lambda=0.1$ & $46.97$ & $64504$  &  $94.89$ &  $8351$ \\ 

 &   & $\lambda=100$ & $52.89$ & $3655$  & $99.97$ &  $1197.2$ \\ \cline{2-7}

 & ADMM  & $\lambda=0.1$ & $53.40$ & $49782$   & $99.24$  & $7163$\\ 

 &   & $\lambda=100$ & $52.89$ & $37111$  & $99.15$ & $6302.3$  \\ 

\cline{2-7}

 & gOMP  & $\kappa=18$ & $46.41$ & $82$  & $100$ & $34.17$\\ 

 &   & $\kappa=24$  & $\mathbf{56.24}$ & $61$   & $100$  & $39.60$\\ \cline{2-7}

 & BIHT  & $\kappa=18$  & $47.03$ & $1914$  & $91.33$  & $1780.7$\\ 

 &   & $\kappa=24$ & $53.01$ & $164$  & $99.34$  & $178.53$\\ 

\cline{2-7}

 & CoSaMP  & $\kappa=18$ & $47.07$ & $1951$  & $88.76$ & $2295.2$ \\ 

 &  & $\kappa=24$ & $52.85$ & $156$  & $99.08$ & $253.50$\\

\hline
%90\addlinespace
\midrule

\multirow{12}{*}{China ($420\times 140\times154$)}

 & FISTA  & $\lambda=0.1$  & $46.39$ & $550638$  &  $97.39$ & $35157$ \\ 
 
 &   & $\lambda=100$  & $47.04$  & $28962$   & $99.86$ &  $8196.7$\\\cline{2-7}
 
 & ADMM  & $\lambda=0.1$ & $47.85$ & $373229$  & $99.79$ & $22940$ \\ 

 &   & $\lambda=100$  & $47.04$ & $244835$  & $99.97$ & $16526$ \\ 

\cline{2-7}

 & gOMP  & $\kappa=16$  & $43.96$ & $456$  & $100$  & $64.28$\\ 

 &   & $\kappa=22$ & $\mathbf{53.59}$ & $346$  & $100$ & $74.13$\\ \cline{2-7}

 & BIHT  & $\kappa=16$  & $42.48$ & $56675$  & $92.41$ & $8983.5$ \\ 

 &   & $\kappa=22$ & $42.99$ & $274$  & $100$ & $136.80$ \\ 

\cline{2-7}

 & CoSaMP  & $\kappa=16$  & $48.76$ & $16586$  & $89.72$ &  $12241$\\ 

 &   & $\kappa=22$ & $\mathbf{55.39}$ & $474$  & $99.97$ & $1069.7$ \\

\hline

\bottomrule
\end{tabularx}
\end{table*}

\begin{figure}[!htbp]
\centering
	\begin{subfigure}{0.09\textwidth} % width of left subfigure
		\includegraphics[width=\textwidth]{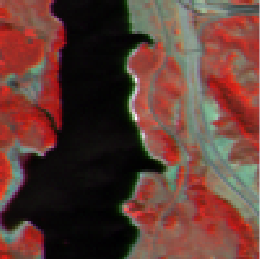}
		\caption{Original} % subcaption
		\label{SubFig:Jasper_original_non_sparisifed}
	\end{subfigure}	
	\begin{subfigure}{0.09\textwidth} % width of left subfigure
		\includegraphics[width=\textwidth]{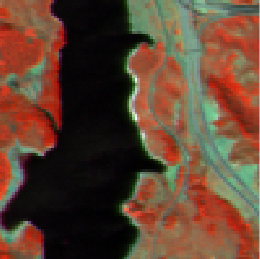}
		\caption{Sparse} % subcaption
		\label{SubFig:Jasper_sparisifed}
	\end{subfigure}	
	\begin{subfigure}{0.09\textwidth} % width of left subfigure
		\includegraphics[width=\textwidth]{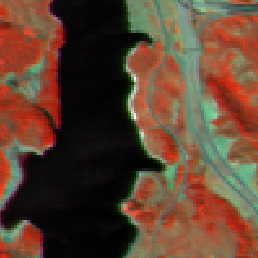}
		\caption{FISTA} % subcaption
		\label{SubFig:Jasper_FISTA_lambda_100}
	\end{subfigure}	
    \begin{subfigure}{0.09\textwidth} % width of left subfigure
		\includegraphics[width=\textwidth]{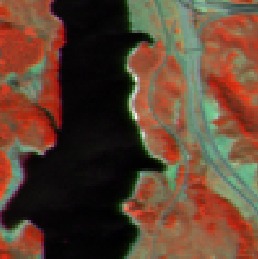}
		\caption{ADMM} % subcaption
		\label{SubFig:Jasper_ADMM_lambda_100_alpha_1_point_8}
	\end{subfigure}	
    \begin{subfigure}{0.09\textwidth} % width of left subfigure
		\includegraphics[width=\textwidth]{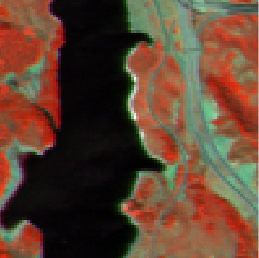}
		\caption{gOMP} % subcaption
		\label{SubFig:Jasper_GOMP_K_24}
	\end{subfigure}	
    \begin{subfigure}{0.09\textwidth} % width of left subfigure
		\includegraphics[width=\textwidth]{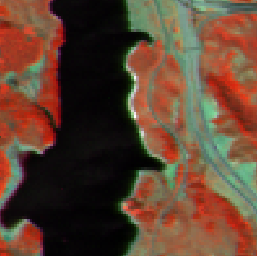}
		\caption{BIHT} % subcaption
		\label{SubFig:Jasper_BIHT_K_24}
	\end{subfigure}		
    \begin{subfigure}{0.09\textwidth} % width of left subfigure
		\includegraphics[width=\textwidth]{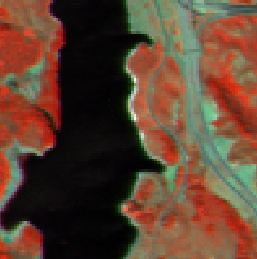}
		\caption{CoSaMP} % subcaption
		\label{SubFig:Jasper_CoSaMP_K_24}
	\end{subfigure}		
       \caption{Jasper Ridge Recoveries}
       \label{Fig:jasper_figures}
\end{figure}

\textit{Salinas}, \textit{Jasper Ridge}, and \textit{China} data sets are used to compare the algorithms. Fig.~\ref{SubFig:Jasper_original_non_sparisifed} shows the false color composite image of the bands 46, 108, and 164 in Jasper Ridge and Fig.~\ref{SubFig:Jasper_sparisifed} presents the image after the sparsification using the empiric factor $T=0.1$ which gives 90.93\% of the samples rounded to zero, and 87.22\% and 90.44\% in Salinas and China, respectively. The \textit{peak signal-to-noise ratio} ($\mathit{PSNR}$) metric is used to measure the quality of a processed image with respect to the original image. The image quality after the sparsification is measured with the $\mathit{PSNR}$ calculated between the sparsified and the original data. For Salinas, Jasper Ridge and China data sets the $\mathit{PSNR}$ is respectively $34.22\;\mathrm{dB}$, $32.22\;\mathrm{dB}$, and  $29.86\;\mathrm{dB}$. After the sparsification, 40\% of the data are randomly subsampled, i.e., $2.5\times$ compression. The following analysis compares, for different empiric arbitrary $\lambda$ and $\kappa$ in addition to the parameters $G$ and $\mu$ experimentally fixed to respectively to $\floor*{\frac{\kappa}{5}}$ and $0.1$, the algorithms in terms of accuracy given by the $\mathit{PSNR}$ calculated henceforth between the recovered and the sparsified data, and the performance through the convergence, the recovery time and the scalability. 
The slow convergence of some the algorithms such as the FISTA/CoSaMP demands that the pixel recovery time is bounded experimentally to $t_{conv.}=2.0\;\mathrm{s}$ for the empirical error tolerance $\epsilon=10^{-8}$ in all the algorithms. 

\subsection{Accuracy: $\mathit{PSNR}$}
Figs.~\ref{SubFig:Jasper_FISTA_lambda_100}~-~\ref{SubFig:Jasper_CoSaMP_K_24} and Table~\ref{summary_table} present the results for the Jasper Ridge data set using $\lambda=100$ and $\kappa=24$. Table~\ref{summary_table} shows that the algorithms reconstruct the data sets with $\mathit{PSNR}$ $\gtrapprox 40\;\mathrm{dB}$. The gOMP and CoSaMP achieve higher $\mathit{PSNR}$ than the other algorithms. Specifically, the CoSaMP algorithm achieves the highest  $\mathit{PSNR}$ in Salinas and China data sets, but the gOMP algorithm reaches higher $\mathit{PSNR}$ than the CoSaMP for the Jasper Ridge. The gOMP algorithm gives similar maximum $\mathit{PSNR}$ values across the data sets in the range of $[53.59-56.24]\;\mathrm{dB}$ whereas the CoSaMP gives a wider range of $[52.85-70.98]\;\mathrm{dB}$. Furthermore, when the sparsity parameters $\lambda$ and $\kappa$ are varied respectively in the range of $[0.1-100]$ and $[16-33]$, the variation of the $\mathit{PSNR}$ across the data sets in average is $0.84\;\mathrm{dB}$, $3.20\;\mathrm{dB}$, $3.31\;\mathrm{dB}$, $10.76\;\mathrm{dB}$, and $14.08\;\mathrm{dB}$, for respectively the ADMM/BIHT/FISTA/gOMP/CoSaMP. Therefore, the gOMP and CoSaMP algorithms despite achieving maximum $\mathit{PSNR}$ present also a larger average variation of the $\mathit{PSNR}$ when different values in $\kappa$ are used.

\subsection{Performance: Convergence, Recovery Time and Scalability} Table~\ref{summary_table} shows that the minimum number of iterations needed to converge in the convex algorithms is three orders larger than in the greedy algorithms. However, the convex algorithms have more mathematical guarantees of convergence than the greedy algorithms. Furthermore, the gOMP algorithm shows a constant order regardless of the value of the parameter $\kappa$ in the three data sets. Additionally, Table~\ref{summary_table} shows the convergence ratio defined as the quantity of pixels that converge under $t_{conv.}$ for the selected tolerance $\epsilon$ with respect to the total number of data cube pixels. Although in most cases increased $\lambda$ and $\kappa$ give higher convergence ratio, Table~\ref{summary_table} shows that the gOMP achieves convergence for all the cube pixels. Regarding the recovery time, Table~\ref{summary_table} shows that the algorithms converge faster when sparser recovery is aimed using higher values in $\lambda$ and lower in $\kappa$. In the convex algorithms, the ADMM with $\alpha=1.8$ achieves a lower recovery time compared to the FISTA, yet the gOMP algorithm achieves the fastest recovery time. Finally, the time scalability is given by the order of complexity of the most demanding operation. In the case of the FISTA/ADMM, the matrix-vector multiplication in Step 1 is the most time-consuming operation with order of complexity $\mathcal{O}(M\times N)$. In the greedy algorithms, the union of the support vectors in Step 1 is the most demanding operation with order of complexity $\mathcal{O}(M+N)$.

%In the greedy algorithms, the gOMP and BIHT make use of the backslash implementation from \textit{MATLAB} to calculate the significant samples which is significantly faster than using the \textit{pinv} implementation from Matlab used in the CoSaMP implementation used in this work. In any case, for the three greedy algorithms, the union of the support vectors in Step 1 is the most demanding operation with order of complexity $\mathcal{O}(M+N)$. 

%% file: 5-Conclusions.tex
\section{Conclusion}
\label{Section: conclusions}

A comparative study of the convex and greedy algorithms for recovery of compressed HSI data is addressed. The algorithms recover three HSI data sets, yet GOMP shows to overperform the other algorithms in terms of the accuracy and performance. In light of these results, further work on the GOMP algorithm is suggested to study how to maximize its accuracy while keeping the performance.

%% file: 6-Acknowledgement.tex
\section*{Acknowledgment}

The research leading to these results has received funding from the NO Grants 2014 – 2021, under Project ELO-Hyp contract no. 24/2020, and the Research Council of Norway grant no. 223254 (AMOS center of excellence).